\documentclass
[authorversion,nonacm,
sigconf = true,     
review = false,     
screen = true,      
anonymous = false,  
]
{acmart}

\AtBeginDocument{%
  \providecommand\BibTeX{{%
    \normalfont B\kern-0.5em{\scshape i\kern-0.25em b}\kern-0.8em\TeX}}}
\usepackage{subcaption}
\usepackage{dsfont}
\usepackage{enumitem}
\usepackage{siunitx} 
\usepackage{mathtools} 
\newcommand{\defeq}{\coloneqq}

\copyrightyear{2022}
\acmYear{2022}
\setcopyright{acmlicensed}\acmConference[GECCO '22]{Genetic and Evolutionary Computation Conference}{July 9--13, 2022}{Boston, MA, USA}
\acmBooktitle{Genetic and Evolutionary Computation Conference (GECCO '22), July 9--13, 2022, Boston, MA, USA}
\acmPrice{15.00}
\acmDOI{10.1145/3512290.3528825}
\acmISBN{978-1-4503-9237-2/22/07}

\begin{document}
\renewcommand{\epsilon}{\varepsilon}
\newcommand{\R}{\mathbb{R}}
\newcommand{\Q}{\mathds{Q}}
\newcommand{\Z}{\mathds{Z}}
\newcommand{\sobol}{Sobol\kern-0.15em\'{ } }
\newcommand{\mli}[1]{\mathit{#1}}
\newcommand{\isep}{\mathinner {\ldotp \ldotp}}

\title{Automated Algorithm Selection for Radar Network Configuration}


\author{Quentin Renau}
\email{renau.quentin@gmail.com}
\affiliation{
  \institution{Thales Research \& Technology, \'Ecole Polytechnique, Institut Polytechnique de Paris}
  \city{Palaiseau}
  \country{France}
}

\author{Johann Dreo}
\email{johann.dreo@pasteur.fr}
\affiliation{
  \institution{CSB lab., Institut Pasteur, Université Paris Cité, Bioinformatics and Biostatistics Hub, Department of Computational Biology, Computational Systems Biomedicine laboratory, F-75015 Paris, France.}
  \city{Paris}
  \country{France}
}

\author{Alain Peres}
\email{alain.peres@thalesgroup.com}
\affiliation{
  \institution{Thales Land \& Air Systems}
  \city{Rungis}
  \country{France}
}

\author{Yann Semet}
\email{yann.semet@thalesgroup.com}
\affiliation{
  \institution{Thales Research \& Technology}
  \city{Paris}
  \country{France}
}

\author{Carola Doerr}
\email{Carola.Doerr@lip6.fr}
\affiliation{
  \institution{Sorbonne Universit\'e, CNRS, LIP6}
  \city{Paris}
  \country{France}
}

\author{Benjamin Doerr}
\affiliation{
  \institution{Laboratoire d'Informatique (LIX), CNRS, \'Ecole Polytechnique, Institut Polytechnique de Paris}
  \city{Palaiseau}
  \country{France}
}

\renewcommand{\shortauthors}{Renau et al.}

\begin{abstract}
The configuration of radar networks is a complex problem that is often performed manually by experts with the help of a simulator. Different numbers and types of radars as well as different locations that the radars shall cover give rise to different instances of the radar configuration problem. The exact modeling of these instances is complex, as the quality of the configurations depends on a large number of parameters, on internal radar processing, and on the terrains on which the radars need to be placed. Classic optimization algorithms can therefore not be applied to this problem, and we rely on ``trial-and-error'' black-box approaches. 

In this paper, we study the performances of 13~black-box optimization algorithms on 153~radar network configuration problem instances. The algorithms perform considerably better than human experts. Their ranking, however, depends on the budget of configurations that can be evaluated and on the elevation profile of the location. We therefore also investigate automated algorithm selection approaches. Our results demonstrate that a pipeline that extracts instance features from the elevation of the terrain performs on par with the classical, far more expensive approach that extracts features from the objective function. 
\end{abstract}




\maketitle

\sloppy

\section{Introduction}
\label{sec:intro}

Radar networks are placed to form detection barriers with the goal to detect incoming objects.
The efficiency of these barriers, i.e., the network coverage or the detection probability has to be optimized taking into account several systemic and environmental constraints such as the areas to defend and exclusion areas in the domain (areas in which radars cannot be placed).

A radar network is composed of several radars.
The radars may differ in type (e.g., with respect to the area or the distance they can sense).
Radars are typically configurable, i.e., they have one or several parameters that determine their behavior. A common parameter of a radar device is the \emph{tilt}, which is the angle between the horizontal plane and the direction of sensing.
Optimizing a radar network therefore comprises the choice of the locations of the radars and their specific configuration.
The number of parameters causes the problem to be intractable, and its modeling shows non-linear constraints, non-convexity and non-differentiability, making it difficult to be analytically described.

Solving the radar network configuration problem therefore relies on \emph{black-box optimization}, where the problem instances do not need to be explicitly modeled but where it suffices that the quality of potential solutions can be assessed through a simulator that evaluates how suitable the suggested configuration is. 

In the literature and in industrial contexts, metaheuristics have been widely investigated and used to solve numerical black-box optimization problems.
Metaheuristics are stochastic methods that do not need a priori information on the function to solve. 
Even if these methods have no optimality guarantee, they have been well performing on different types of problems, from traffic congestion~\cite{BotherSFMK021} to RNA design~\cite{MerleauS21}.

In this paper, we seek to maximize the coverage of a terrain using a radar network.
The network is composed of four radars of two types that differ from their parameters.

In order to find the locations and configuration of the radars, we compare the performance of $13$ metaheuristics on $153$ problem instances.
We also investigate the performance of an automated algorithm selection procedure by building two selectors.
One selector follows the traditional approach and is built using landscape feature extracted on the radar objective function.
The second selector is built extracting landscape features on the physical landscape only.

In this paper, we exhibit the complementarity of metaheuristics performances on the radar network configuration problem.
We observe that different algorithms perform well on different instances.
We also observe that different algorithms perform well on different budget of function evaluations.

We also compare the two selectors performances and find that they have similar performances while one is computationally much cheaper to build than the other.

\textbf{Paper organization:} The radar configuration problem is introduced in Sec.~\ref{sec:use-case}. Specifics about our experimental setup are provided in Sec.~\ref{sec:expe}, and the results of our experiments are presented in Sec.~\ref{sec:perf}. The comparison with the human expert is made in Sec.~\ref{sec:hand}.  
Finally, Sec.~\ref{sec:auto} presents the two automated algorithm selection pipelines and analyzes their performance. Our paper is concluded in Sec.~\ref{sec:conclu} with directions for future work.    

\textbf{Reproducibility:} 
While we cannot make the implementation of the radar configuration problem available in open-access format, we provide binaries of the radar network configuration use-case~\cite{dataRadar}. 
It contains the objective function presented in this paper, thus it is possible to recreate the $153$ instances presented in Sec.~\ref{sec:instances} and to generate any instance on any terrain. 
The detailed performance of algorithms for $500$ and $2{,}500$ function evaluations are available at~\cite{dataRadar}. They follow the data format of the IOHprofiler~\cite{IOHprofiler} environment and can hence be easily analyzed and visualized by the IOHanalyzer module~\cite{IOHanalyzer} of the IOHprofiler project~\cite{IOHprofiler}, available online at \url{https://iohanalyzer.liacs.nl/}. Some of the plots in this paper were also created with IOHanalyzer.

\section{Problem Description}
\label{sec:use-case}

Radar configuration problems differ with respect to the object to detect, the number and types of the radars that can be placed, their parameters, the terrain they shall protect, and with respect to constraints on the locations in which the radars can be placed. We briefly describe these aspects in the following paragraphs.

\textbf{Object Characteristics. }
For the sake of simplicity, we consider a simple object model to be detected by the radars.
We suppose the speed of the object to be constant. 
The object altitude above ground level $h$ is also supposed to be constant, i.e., if the ground altitude above sea level is $z$, then the object is flying above the ground at altitude $z+h$.

We define the object angle $\theta$ as the angle between the object direction and the North. This angle is called the \emph{azimuth}.

\textbf{Terrain Definition. }
\label{sec:terrain}
The terrain to cover is a square of $\SI{50}{\kilo\meter} \times \SI{50}{\kilo\meter}$ and for each point, the object can face in any direction.
Therefore, the domain $\Delta$ to cover is represented by voxels $(x,y,\theta)$, where $\theta$ is the object azimuth. For our model, we split each axis of $\Delta$ into $30$ equal parts, resulting in a total number of $30^3 = 27{,}000$ voxels that need to be protected.

The geographical data used to model the terrain is coming from the NASA Shuttle Radar Topography Mission~\cite{srtm_nasa} (SRTM).
These digital elevation models (DEM) cover the entire globe and are provided in mosaics of 5 degrees by 5 degrees tiles.
In order to use complete data, we used DEM that were post-processed with interpolation in~\cite{srtm} (Data available at \url{https://srtm.csi.cgiar.org}).

\textbf{Radar Network Modeling. }
\label{sec:network}
We consider four radars $R$ of two types: one rotating and three staring radars.
These two types only differ by the angular domain that the radar can sense.
In our simulations, we assume that radars with rotating antennas can sense all around them at any time, whereas radars with staring antennas can only sense predefined regions around their \emph{staring direction}.

The number of tunable parameters depends on the type of the radar: the staring radars have four parameters that can be modified: 
\begin{enumerate}
    \item the $x$ ($\approx$ longitude) location in the terrain;
    \item the $y$ ($\approx$ latitude) location in the terrain;
    \item the \emph{tilt}, i.e., the angle between the horizontal plane and the direction of sensing;
    \item the \emph{staring angle}, i.e., the direction the radar will look to.
\end{enumerate}
For rotating radars, the staring angle is not needed and we therefore have three parameters to tune.
The dimension of our radar configuration problem is thus $d= 3+ 3\times 4 = 15$. 

After normalization, our decision space is modeled as $[0,1]^{15}$, i.e., each choice $\xi$ of radar location and configuration (\emph{network} in the following), is a 15-dimensional vector with entries in $[0,1]$. 

Detecting is not a situation where the object is either detected or not detected.
The detection relies on a performance indicator which is called the \emph{probability of detection}.
Computing the probability of detection of one radar is very complex and depends on the radar type and its processing along as the object to detect. 

Multiple radars are gathered into a \emph{network}.
A network is used to aggregate the probabilities of detection for each radars.
In this paper, the probability of detection of the network is given by multiplying the probabilities of non-detection for each radar and returning the corresponding probability of detection.
We assume that the radars are independent and compute:
$$ P_d^{\xi}(x,y,\theta) \defeq 1 - \prod_{i=1}^4  \left(1-P_d^{R_i}(x,y,\theta)\right),$$
where $P_d^{R_i}(x,y,\theta)$ is the probability of detection of the radar $R_i$ given its location $x,y$ and the azimuth $\theta$ of the object.

\textbf{Objective Function. }
\label{sec:objective}
The objective function aims at maximizing the coverage of $\Delta$.
For a given problem instance (i.e., for a given area $T$ to cover), our objective is to find the location and the configuration of the radars such that the number of covered voxels is as large as possible. 
For a given radar network $\xi$, we consider a voxel $(x,y,\theta)$ covered if the probability of detection $P_d^{\xi}(x,y,\theta)$ is greater than or equal to a predefined fixed threshold $\tau$. 

For a given problem instance $T$, the objective function for the coverage scenario is then defined as 
$f_T : [0,1]^{15} \to [0\isep27{,}000]$:
$$f_T(\xi) \defeq \sum\limits_{(x,y,\theta) \in [1\isep30]^3}\mathds{1}\left(P_d^{\xi}(x,y,\theta) \geq \tau \right).$$

Our objective is thus to find the optimal radar network $\xi^*$ maximizing $f_T(\xi)$: 
\begin{equation*}
    \label{eq:coverage}
    \xi^* = \arg\max_{\xi \in \Xi} f_T(\xi).
\end{equation*}

\section{Design of Experiment}
\label{sec:expe}

In this section, we briefly describe the 153 problems instances that we selected for our experiments and the $13$ algorithms that we used to solve these instances.

\subsection{Problem Instances}
\label{sec:instances}
Each instance is defined by an individual terrain (see Sec.~\ref{sec:terrain}).
Overall, 17 different tiles of the digital elevation model (DEM) from all around the world were selected manually, so as to represent a large variety of terrains. 

The size of our use-case (see Sec.~\ref{sec:terrain}) is smaller than the area of the DEM tiles.
To match the dimensions of our use-case, we therefore subsampled the original tiles domains.
As there is no particular reason to favor one area over another, we applied the same downsampling mask to each DEM tile. The mask was created by choosing the upper left $\SI{50}{\kilo\meter}\times\SI{50}{\kilo\meter}$ square and by sampling eight additional squares of the same area uniformly at random, obeying to the condition that the areas shall be pairwise disjoint (i.e., non-overlapping). This mask provides us with nine subsamples for each terrain.
The result is a total number of $9 \times 17 = 153$ instances for our radar network configuration problem.

The instances are labeled by the region name of the DEM tile and by the number of its sample in the mask, i.e., the first sample in the mask for the Brasil region has the instance name \emph{brasil0}.
Each problem instance was generated using the objective function available in~\cite{dataRadar}.

The difference between the highest and the lowest points in each instance is reported in Tab.~\ref{tab:elevation}.
This information gives a hint of the actual landscape of the terrain.
It also gives a first idea about which problems might be easier to solve than others, as, intuitively, flat areas should be easier to cover than mountainous ones.

We define \emph{flat instances} to be instances where the difference between the highest and lowest point is below $100$ meters. 
\emph{Mountainous instances} have a difference between the highest and lowest points greater than $1{,}000$ meters. We have a total number of 36 flat instances, 60 mountainous instances, and 57 intermediate ones. 

\begin{table}[t]
\centering
\caption{Difference between the highest and lowest altitudes (in meters) of each instance. Columns represent the subsample in the DEM tile and rows corresponds the region of the DEM tile.}
\label{tab:elevation}
\footnotesize
\begin{tabular}{l|rrrrrrrrr}
            & \multicolumn{9}{c}{Instance}                                \\
            & 0    & 1    & 2    & 3    & 4    & 5    & 6    & 7    & 8    \\
\hline
afghan. & 1616 & 1266 & 1152 & 1161 & 1534 & 1177 & 1085 & 1144 & 1023 \\
argentina   & 24   & 24   & 24   & 24   & 27   & 22   & 26   & 26   & 27   \\
australia   & 189  & 181  & 117  & 67   & 376  & 93   & 203  & 384  & 90   \\
belarus     & 166  & 196  & 145  & 158  & 135  & 187  & 154  & 166  & 147  \\
brasil      & 61   & 79   & 85   & 108  & 89   & 90   & 129  & 58   & 107  \\
canada      & 1494 & 1571 & 1419 & 1452 & 1528 & 1194 & 1474 & 1515 & 1387 \\
chile       & 1408 & 1311 & 1555 & 2625 & 1252 & 2469 & 1869 & 1330 & 2523 \\
china       & 1417 & 1439 & 1462 & 893  & 1398 & 1039 & 924  & 1271 & 956  \\
congo       & 66   & 70   & 68   & 84   & 64   & 74   & 88   & 64   & 87   \\
france      & 161  & 329  & 300  & 325  & 331  & 177  & 189  & 186  & 307  \\
india       & 3828 & 2111 & 3821 & 2224 & 1983 & 3195 & 3175 & 2070 & 3755 \\
iran        & 914  & 487  & 841  & 929  & 788  & 788  & 986  & 319  & 979  \\
moldavia    & 111  & 245  & 241  & 235  & 257  & 166  & 170  & 187  & 229  \\
nepal       & 1439 & 1753 & 1232 & 1656 & 1723 & 1367 & 1529 & 1644 & 1264 \\
russia      & 341  & 302  & 341  & 343  & 302  & 327  & 355  & 330  & 322  \\
sahara      & 25   & 25   & 24   & 43   & 20   & 28   & 25   & 27   & 29   \\
usa         & 1161 &  1486 &  1518 &  1300 & 1321 &  1049 & 1173 & 1393 & 1228
\end{tabular}
\end{table}

\subsection{Algorithm Portfolio}
\label{sec:portfolio}
The portfolio of algorithms is composed of 12 of the algorithms with their default implementation parameters:
\begin{itemize}[leftmargin=*]
    \item Differential Evolution~\cite{StornP97} with \emph{scipy}~\citep{scipy} (version 1.5.4) implementation;
    \item Nelder-Mead~\cite{NelderMead} with \emph{scipy} implementation;
    \item Powell's method~\cite{powell} with \emph{scipy} implementation;
    \item L-BFGS-B~\cite{bfgs} with \emph{scipy} implementation and a finite differences step of $0.01$;
    \item Particle Swarm Optimization (PSO)~\cite{KennedyPSO95} with global best topology and  \emph{pyswarms}~\cite{pyswarms} (version 1.3.0) implementation; 
    \item Random Search with uniform sampling;
    \item Quasi-Random Search with \sobol~\cite{sobol_distribution_1967} low-discrepancy sequences;
    \item Five CMA-ES variants from the ModCMA framework~\cite{Rijn0LB16}: 
    \begin{itemize}
        \item Vanilla CMA-ES~\cite{HansenO01} (configuration 00000000000 in the ModCMA framework);
        \item CMA-ES with mirrored sampling~\cite{BrockhoffAHAH10} and pairwise selection~\cite{AugerBH11} (configuration 00100001000);
        \item CMA-ES with elitism (configuration 01000000000);
        \item CMA-ES with active update~\cite{JastrebskiA06} (configuration 10000000000);
        \item CMA-ES with active update, elitism and BIPOP increasing population size~\cite{Hansen09} (configuration 11000000002).
    \end{itemize}
\end{itemize}

We expand our portfolio of algorithms by performing algorithm configuration on Differential Evolution on one mountainous instance, the \emph{chile0} instance, in order to create specific solvers. 
To this end, we use the \emph{irace}~\cite{irace} (version 3.4.1) configurator with default parameters and the lowest budget possible of runs: $180$.
This choice of number of runs is motivated by the expensive computation time of the objective function. 
Running irace on this single use-case takes around $24$ hours on our computer hardware. 
We also tune other algorithms on single or multiple instances, but the results of tuned algorithms were comparable to their default configuration.

The parameters of Differential Evolution to tune are the population size (between $4$ and $100$), the mutation (between $0$ and $2$) and, recombination constants (between $0$ and $1$). 
Compared to the default setting, irace suggests a much smaller population size and a smaller mutation constant (see Tab.~\ref{tab:de_param}).
The tuned version of the DE algorithm enters our portfolio of algorithms under the name DE\_2500\_chile.

\begin{table}[t]
\centering
\caption{Differential Evolution population size, mutation and recombination parameters. Comparison of default algorithm parameters with tuned algorithm parameters.}
\label{tab:de_param}
\begin{tabular}{l|ll}
              & DE (default)      & DE\_2500\_chile \\
              \hline
popsize       & 15       & 6               \\
mutation      & $\mathcal{U}$(0.5,1) & 0.1             \\
recombination & 0.7      & 0.58           
\end{tabular}
\end{table}
\raggedbottom

\subsection{Experimental Setup}
\label{sec:setup}
We run each algorithm $30$ independent times on each problem instance. For each of these runs, we log the whole performance trajectory, i.e, we log all the improvements achieved by the algorithms up to a budget of 2{,}500 function evaluations. For the analyses presented in this paper, we mainly focus on one \textbf{low-budget setting with 500 function evaluations} and one \textbf{large-budget setting with 2{,}500 function evaluations}.

As most default optimization algorithm implementations assume minimization, we seek to find a network configuration that minimizes the non-covered areas.

\section{Algorithm Performances}
\label{sec:perf}
In this section, we discuss the performances of the algorithms from our portfolio when applied to the use-case defined in Sec.~\ref{sec:use-case}. We analyze the performance distribution over the 30 runs per instance in Sec.~\ref{sec:run} and regard the median performance over time in Sec.~\ref{sec:time}. A statistical comparison in Sec.~\ref{sec:stat} underlines that these runtime distributions are significantly different, which in principle suggests to use algorithm configuration and selection.
In Sec.~\ref{sec:sbs_unconstrained}, we give the performance of the different algorithms aggregated over all 30 runs on all 153 instances and derive from this a single best solver. 

\subsection{Individual Runs}
\label{sec:run}

\begin{figure}
    \centering
     \begin{subfigure}[htbp]{0.45\textwidth}
         \caption{\emph{sahara0}}
         \label{fig:sahara_500}
         \centering
         \includegraphics[trim={0 4cm 0 0},clip,width=\textwidth]{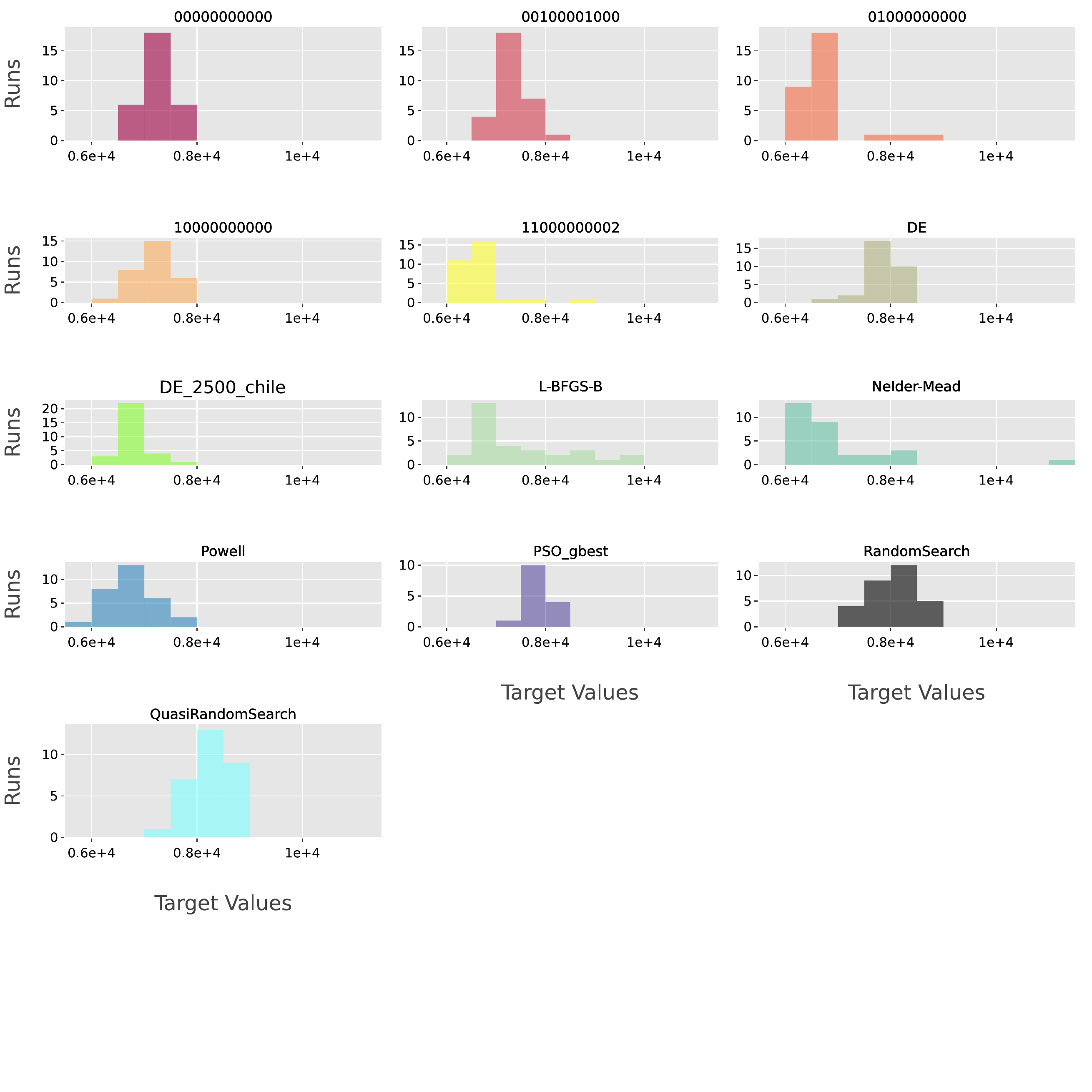}
     \end{subfigure}
     \hfill
     \begin{subfigure}[htbp]{0.45\textwidth}
          \caption{\emph{nepal3}}
         \label{fig:nepal_500}
        \centering
         \includegraphics[trim={0 4cm 0 0},clip,width=\textwidth]{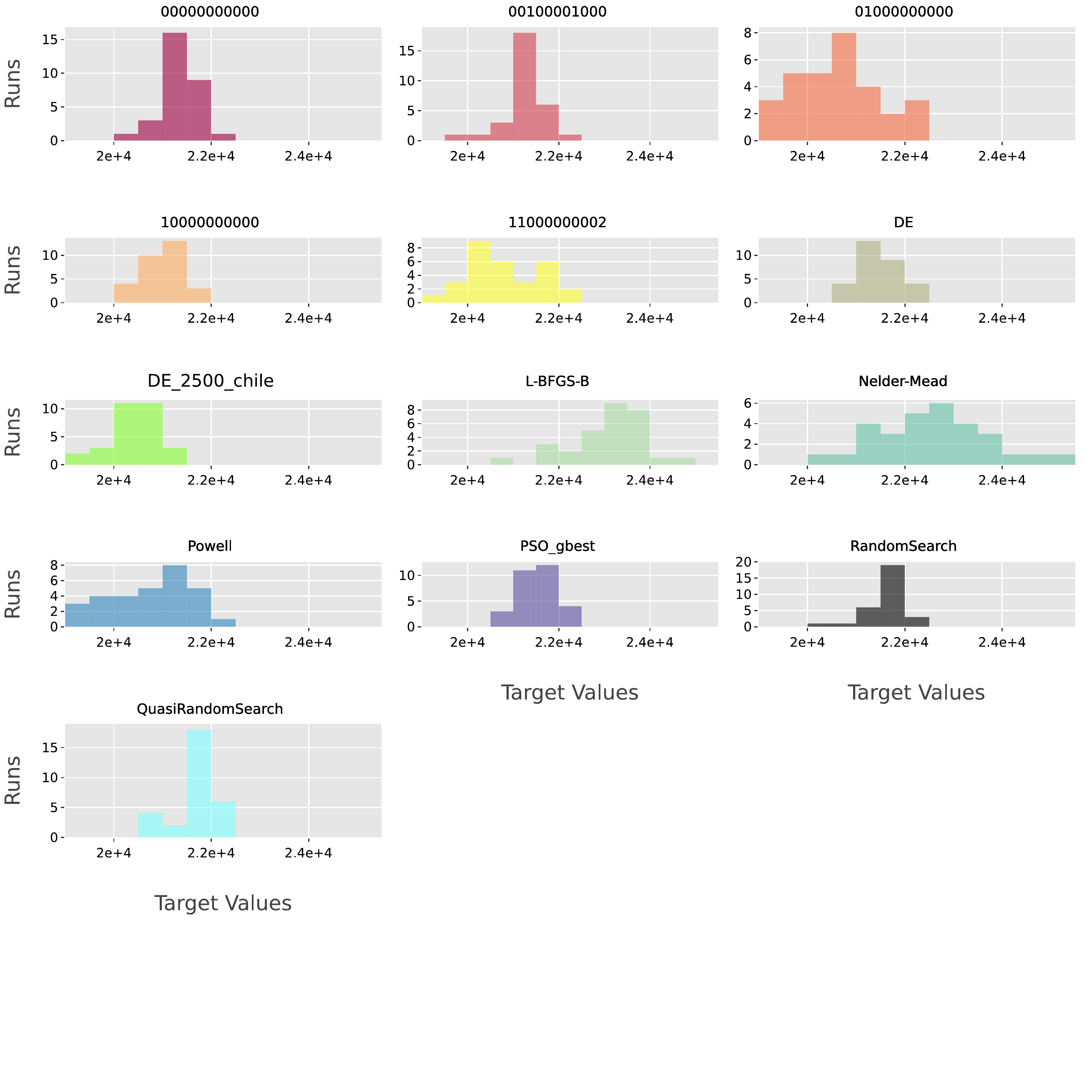}
     \end{subfigure}
     \caption{Histograms (over 30 runs) of the objective values reached on the unconstrained use-case in $500$ function evaluations. }
     \label{fig:histo_500}
\end{figure}

Figures~\ref{fig:histo_500} show the histograms of the objective values reached in the 30 runs of the 13 algorithms on two terrains and using the small budget of 500 evaluations. Within a terrain type, the scaling of the $x$ axis and the bins used in the histograms are identical each algorithm; they differ between the terrain types because of the significantly different objective values obtained -- naturally, the radar network configuration problem is much harder in the rugged \emph{nepal3} terrain than in the flat \emph{sahara0} terrain.

At a first glance, all algorithms have a reasonable performance. The simple approaches of random and quasirandom search, mostly given for comparison, naturally are often beaten by the more advanced algorithms, in particular, on the less rugged terrain.

On both instance, Powell's method found the best solution most often. However, given the dispersion of solution values, Powell's method is not the best performing algorithm in median on either instance. Here, some DE variant, Nelder-Mead (on the \emph{sahara0} instance) or, DE\_2500\_chile are superior. 

\subsection{Influence of the Budget}\label{sec:time}

\begin{figure}
    \centering
     \begin{subfigure}[htbp]{0.233\textwidth}
         \caption{\emph{sahara0}}
         \label{fig:sahara_convergence}
         \centering
         \includegraphics[width=\textwidth]{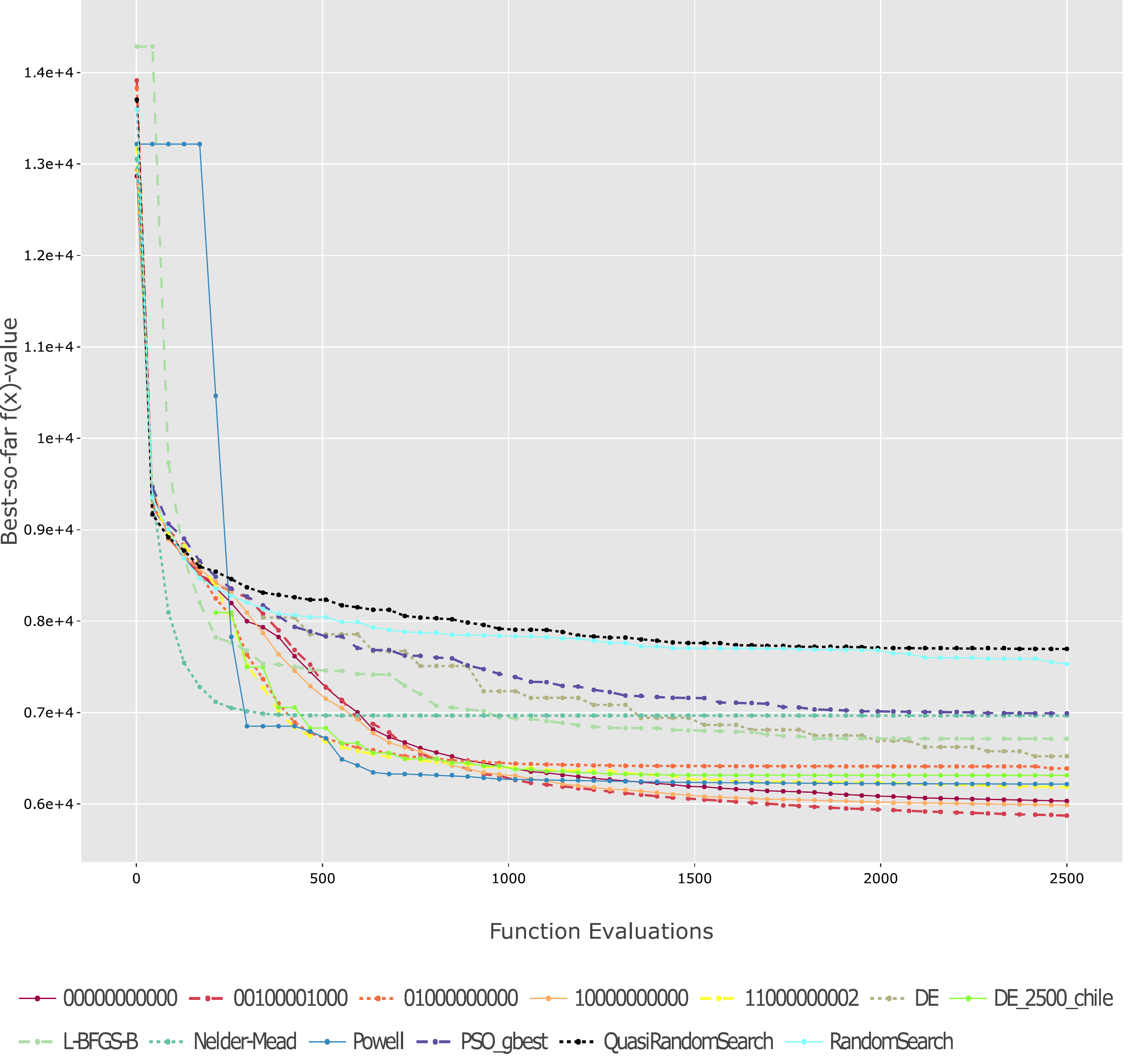}
     \end{subfigure}
     \hfill
     \begin{subfigure}[htbp]{0.233\textwidth}
         \caption{\emph{nepal3}}
         \label{fig:nepal_convergence}
         \centering
         \includegraphics[width=\textwidth]{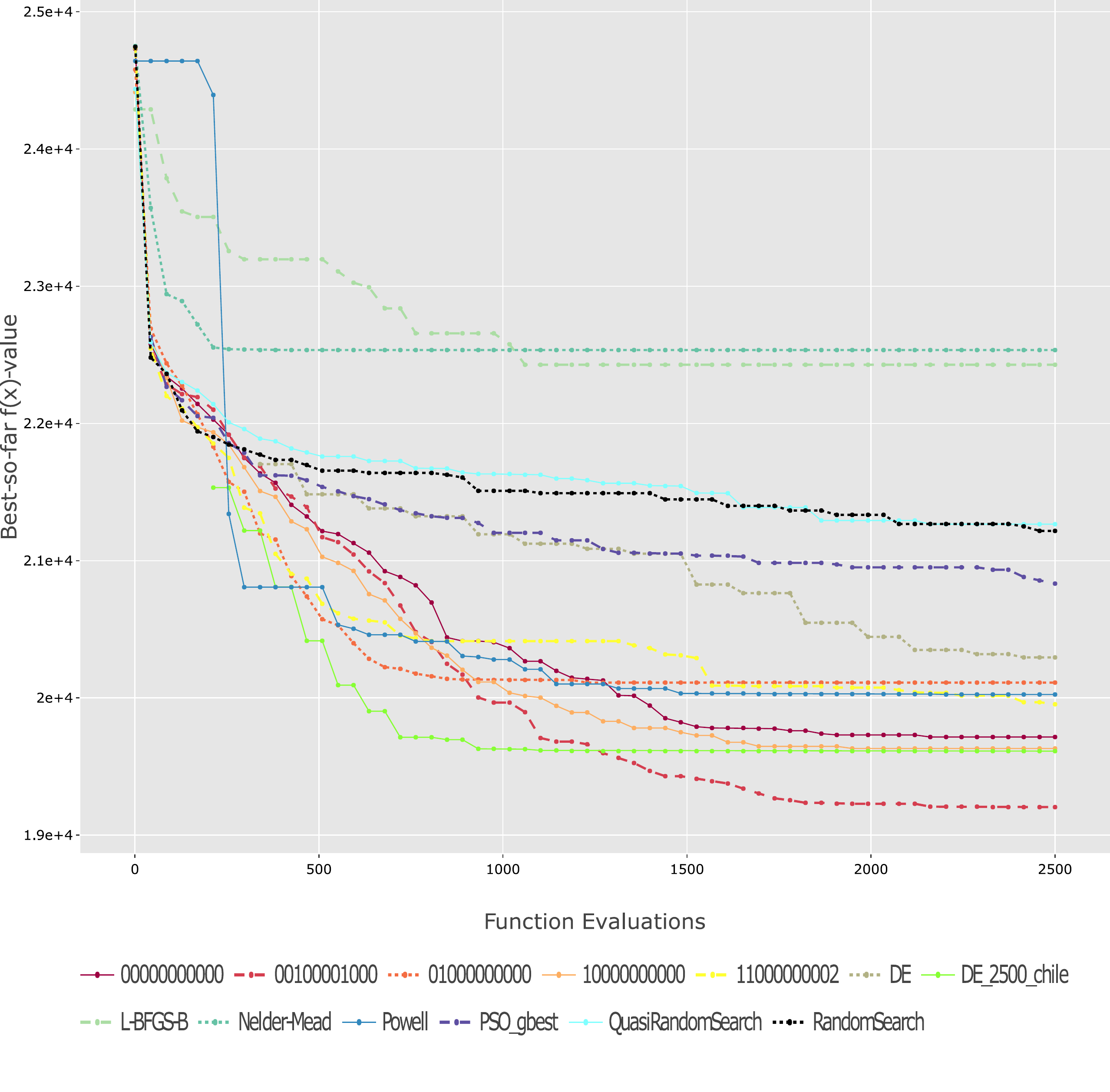}
     \end{subfigure}
     \caption{Median objective values over time, for 30 runs, for \emph{sahara0} and \emph{nepal3} instances, for the 13 algorithms of our portfolio.}
     \label{fig:convergence}
\end{figure}

The relative performance of the algorithms not only depends on the instance type, but also crucially on the computational budget. This is visible from Fig.~\ref{fig:convergence}, showing for each algorithm how the median objective values out of $30$ runs develops over time. On both instances, the CMA-ES variant \emph{00100001000} performs very well from around $1{,}000$ function evaluations on, whereas for smaller budgets other algorithms are significantly better, e.g., DE\_2500\_chile.

The differences between the algorithms are less pronounced for the \emph{sahara0} instance, i.e., around a $300$ voxels difference between the best and 5-th best algorithm after $2{,}500$ iterations over a domain size of 27,000 voxels. For the more rugged \emph{nepal3} landscape, however, the differences are noteworthy as, apparently, the different algorithms get stuck in local optima of very different solution quality.

\textbf{Budget and Elitism. }
The results presented in Sec.~\ref{sec:time} seem to show that some CMA-ES variants are preferable for small budgets whereas others perform well on large budgets.

The main differences between these variants are their respective selection mechanism.
Variants that perform well for the low budget are elitist, i.e., they use plus-selection, whereas variants that perform well for the larger budget are non-elitist, i.e., they use comma-selection.

\begin{figure}
    \centering
     \begin{subfigure}[htbp]{0.2\textwidth}
         \caption{\emph{sahara0}}
         \label{fig:cma_sahara}
         \centering
         \includegraphics[width=\textwidth]{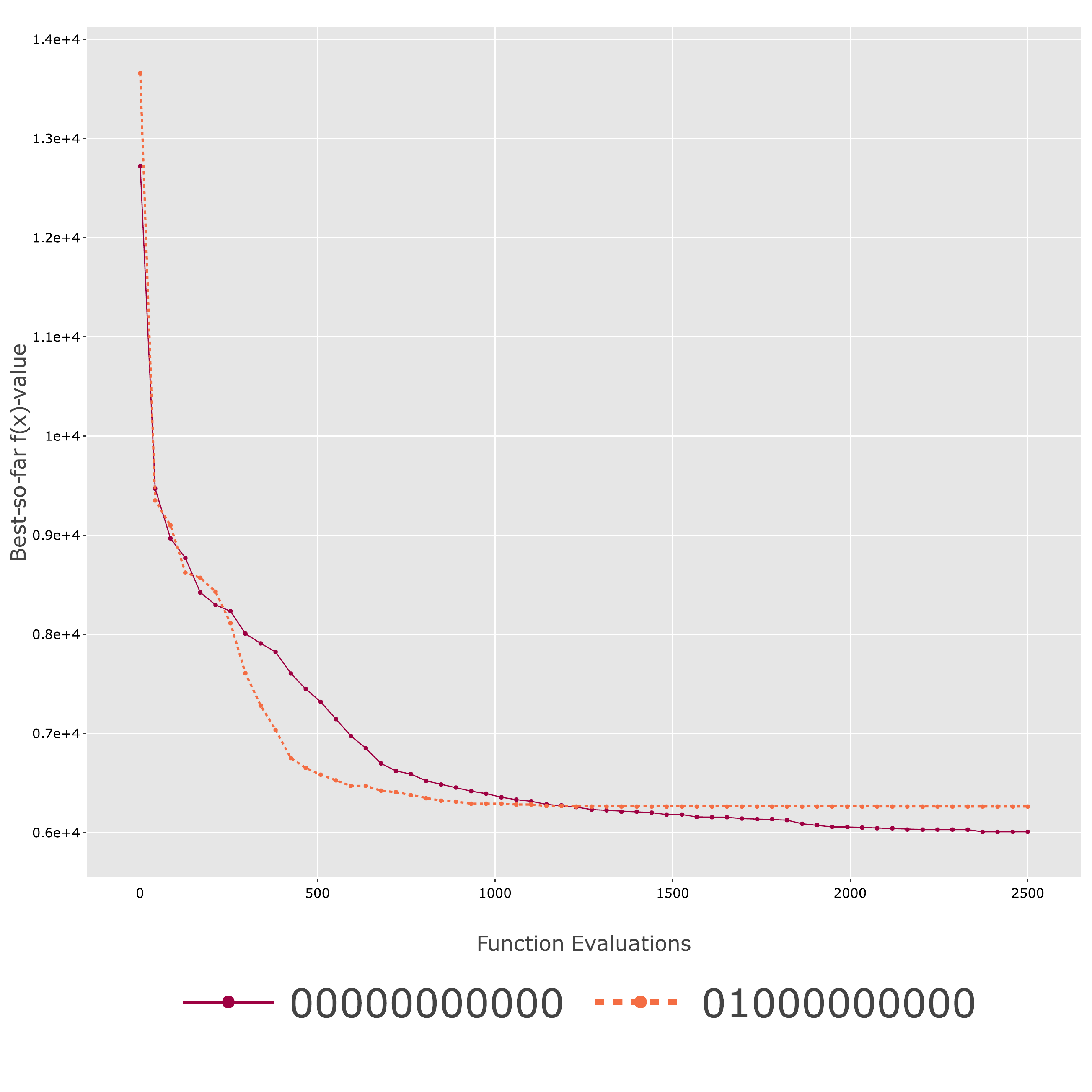}
     \end{subfigure}
     \hfill
     \begin{subfigure}[htbp]{0.2\textwidth}
         \caption{\emph{nepal3}}
          \label{fig:cma_nepal}
         \centering
         \includegraphics[width=\textwidth]{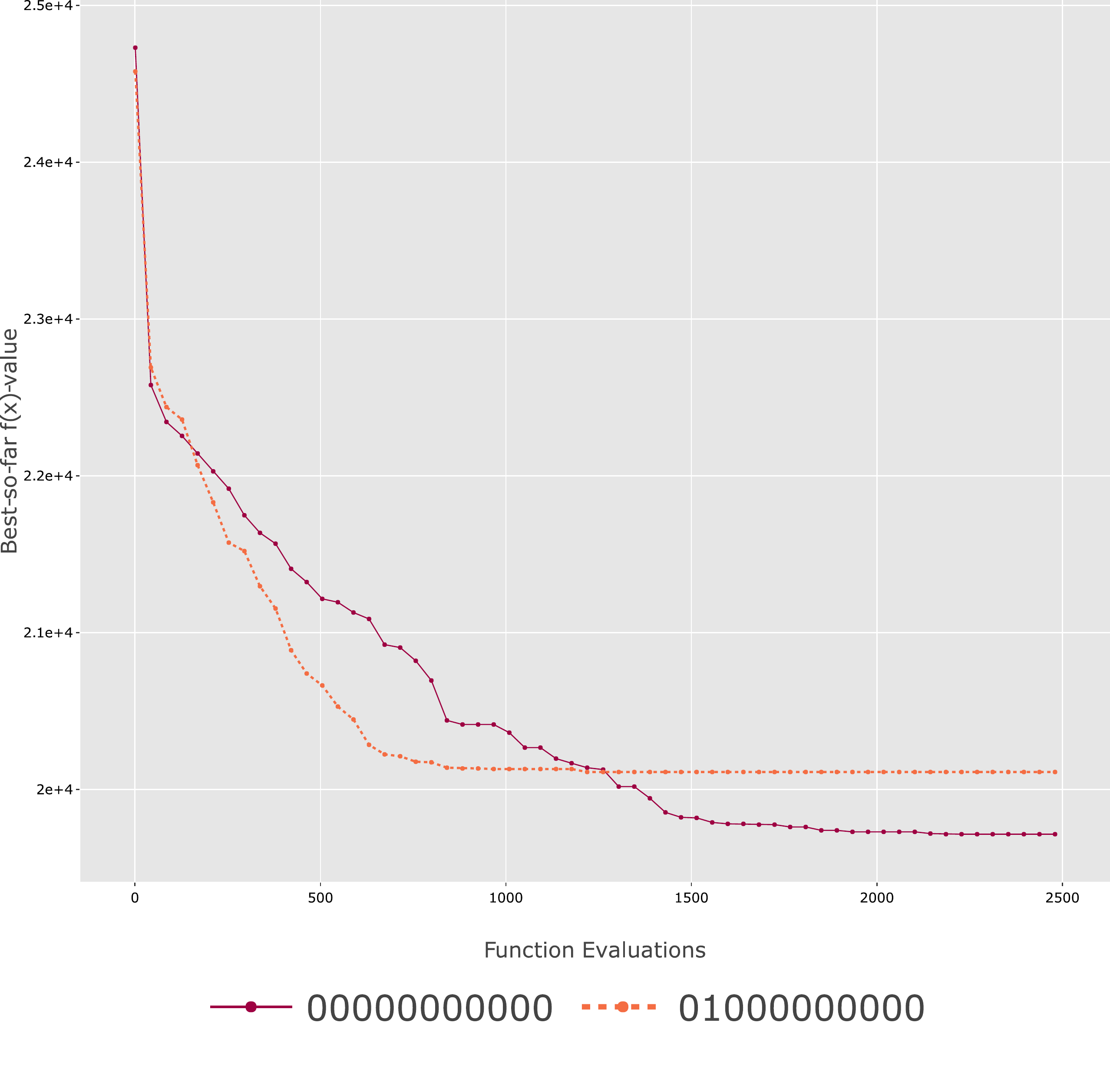}
      \end{subfigure}
     \caption{Median objective values over time, for 30 runs of vanilla (red) and elitist (orange) CMA-ES variants with different selection mechanism.}
     \label{fig:cma}
\end{figure}

Fig.~\ref{fig:cma} illustrates this behavior on the \emph{sahara0} (Fig.~\ref{fig:cma_sahara}) and the \emph{nepal3} instances (Fig.~\ref{fig:cma_nepal}) of two CMA-ES variants.
One variant is the non-elitist vanilla CMA-ES (00000000000) and the other variant uses the elitism module (0100000000).
On both instances, we observe that the elitist variant performs better on lower budgets.
In contrast, the non-elitist variant is performing better for larger budgets.
On all instances, there is a number of function evaluations where the non-elitist performance curve crosses the elitist one.

The crossing point occurs around the $900$ evaluations on the \emph{sahara0} and around $1{,}300$ evaluations on \emph{nepal3}.
Overall, this crossing occurs on all instances between $600$ and $1{,}300$ evaluations.
Nevertheless, the crossing seems not to depend of the landscape of the terrain, i.e., the variants performances may cross each other later on some flat instances than mountainous instances and vice versa.
As an example, the crossing occurs around $1{,}300$ evaluations on the \emph{brasil1} instance which is rather flat and around $750$ evaluations on the \emph{chile6} instance.

\subsection{Statistical Significance}
\label{sec:stat}
All algorithms seem to have similar performances.
To confirm this behavior, we perform a Kolmogorov-Smirnov test to determine if one algorithm outperforms all others on each instance.
We perform this test for each pair of algorithms with a confidence level $\alpha=0.01$.

In the \emph{nepal3} instance, for a budget of $500$ function evaluations, the best performing algorithm, \emph{DE\_2500\_chile}, is statistically better than the majority of algorithms except for \emph{Powell}, and the \emph{11000000002} and \emph{01000000000} CMA-ES variants.

For a larger budget of $2{,}500$ function evaluation, \emph{Powell}, \emph{DE\_2500\_chile} and, all CMA-ES variants perform better than other algorithms but are do not show statistically different performances between them.

This tendency of multiple algorithms having similar performances can also be seen for \emph{sahara0} instance and all other instances in the benchmark (see~\cite{dataRadar})

\subsection{Single Best Solvers (SBS)}
\label{sec:sbs_unconstrained}

The results of the previous subsections show that the different algorithms have substantially different strengths. This is what motivated us to use an automated algorithm selection approach. Before doing so, we nevertheless try to derive some insight on which of the algorithms in some general sense are superior.

For this comparison, we again regard the median performance of an algorithm on an instance. This defines, for each instance and each of the two budgets a best-performing algorithm.

\begin{figure}[t]
    \centering
     \begin{subfigure}[htbp]{0.49\textwidth}
          \caption{$500$ function evaluations}
         \label{fig:cummulative_500}
        \centering
         \includegraphics[width=\textwidth]{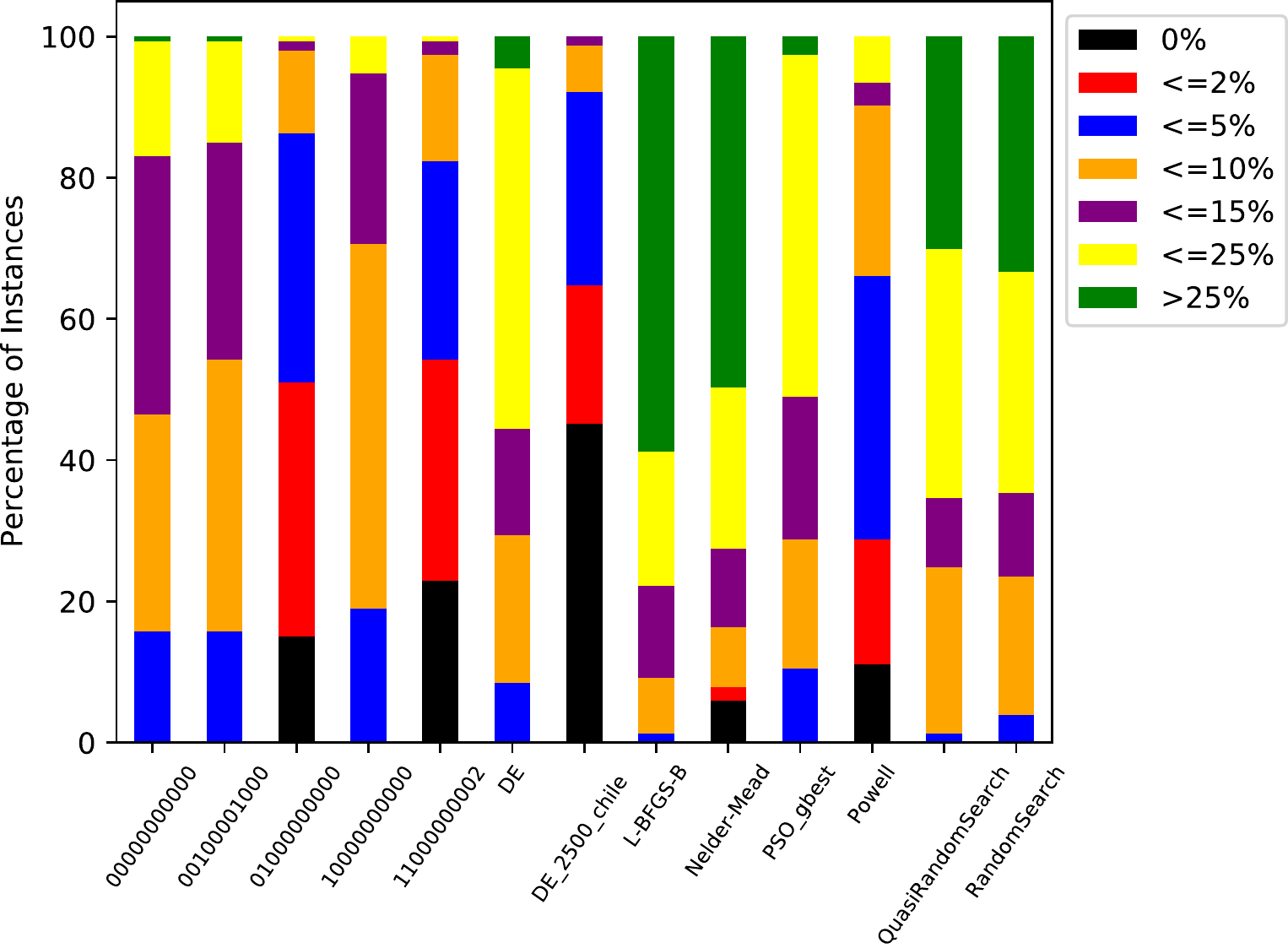}
     \end{subfigure}
     \hfill
     \begin{subfigure}[htbp]{0.49\textwidth}
         \caption{$2{,}500$ function evaluations}
         \label{fig:cummulative_2500}
         \centering
         \includegraphics[width=\textwidth]{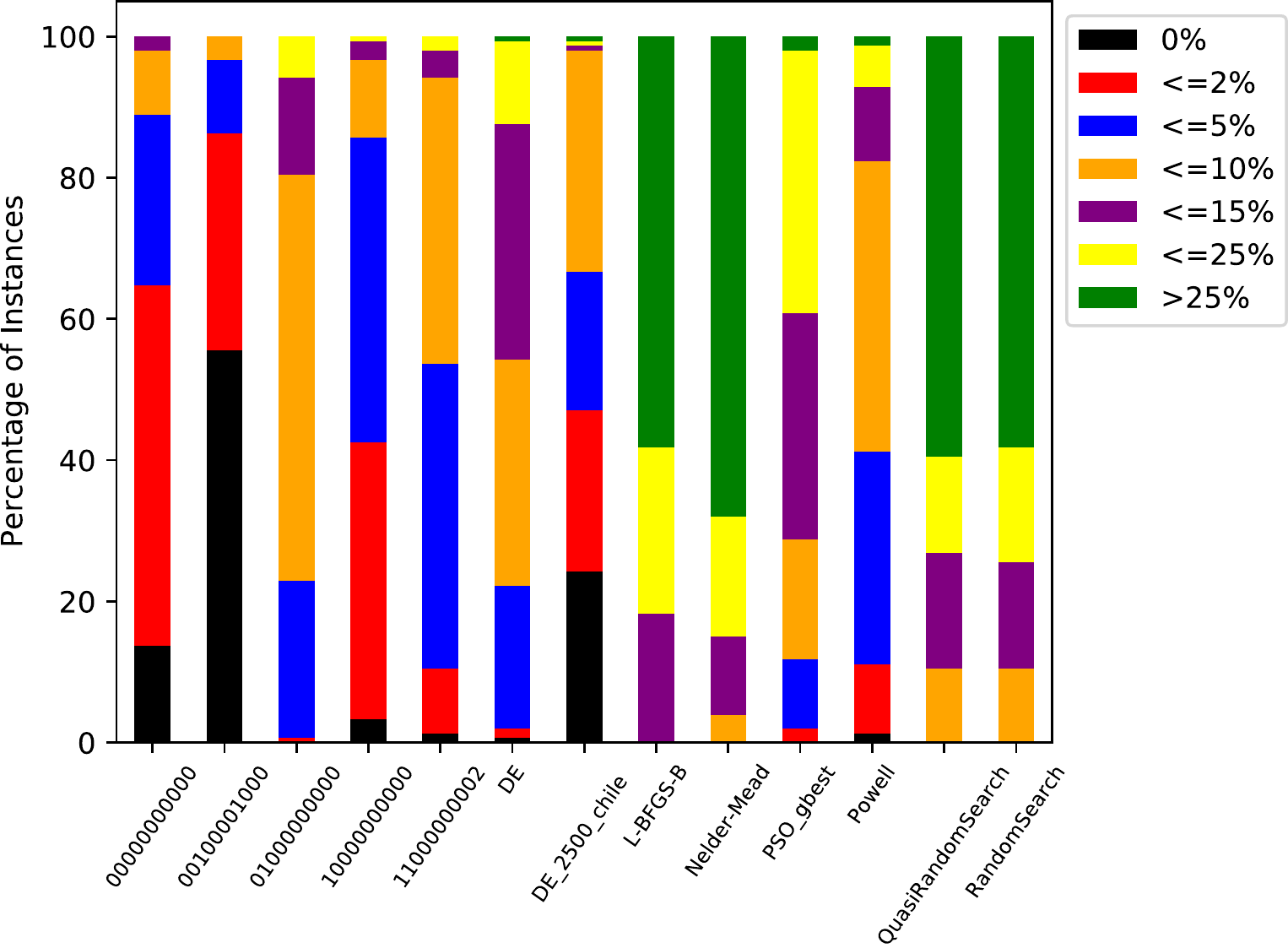}
     \end{subfigure}
     \caption{Percentages of instances solved within $X\%$ of the best algorithm performance.}
     \label{fig:cummulative}
\end{figure}

We then compare each algorithm on each instance with the performance of this best-performing algorithm. Fig.~\ref{fig:cummulative} displays the distribution of the solution quality loss (in percent) of each algorithm. We see, for example, that for a computational budget of 500, DE\_2500\_chile obtained the best result on 46\% of the instances (that is, on 69 out of 153 instances), and that it never encountered a loss of more than 15\%. Other generally good algorithms for this budget are the CMA-EA variants \emph{11000000002} and \emph{01000000000} as well as \emph{Powell's method}.

To define a \emph{single best solver}, denoted by $\mli{SBS}_{500}$ and $\mli{SBS}_{2{,}500}$, we compute the median (over 153 instances) performance loss of each algorithm. This loss corresponds to the difference between the performance of the best algorithm on a given instance with the performance of the target algorithm.
For the budget of 500 evaluations, this number is smallest for DE\_2500\_chile, namely as little as $0.39$\%. For a budget of $2{,}500$, apparently the ES variant \emph{00100001000} with a median performance loss of $0$\% is the $\mli{SBS}_{2{,}500}$. This algorithm performed best on $86$ of the $153$ instances.

\section{Comparison with Manual Optimization}
\label{sec:hand}

In practice, the configuration of radar networks is often hand-designed by experts with the help of a simulator.
The objective of this section is to compare hand-designed solutions with those of the 13 algorithms from our portfolio.

\textbf{Problem Instance.} 
The problem considered to be solved by hand is the \emph{canada0} instance. 
The landscape can be divided into two areas: a rather flat area on the left and a peaked area with mountains on the right.

\textbf{Algorithm Performances on \emph{canada0}.} 
The best performing algorithm on this problem is vanilla CMA-ES (00000000000) with a median number of $15{,}169$ voxels that are not covered and a best run of $14{,}125$ voxels not covered for $2{,}500$ function evaluations.

For a smaller budget of $500$ function evaluations, the CMA-ES variant $11000000002$ achieves the best median performance with a fitness of $16{,}518$.
Powell's method performs the best run with a value of $15{,}841$.

For comparison, the median value of random search for the large budget is $17{,}865$ and it is $18{,}545$ for the small budget.

\textbf{Radar Network Configuration Contest}. 
In order to gather a sufficient amount of data to compare hand-designed solutions to algorithms, we launched a contest to solve the \emph{canada0} instance. The competition had 13 participants of diverse expertise in radar network configuration, ranging from engineers with significant radar \& optimization expertise, to first-year PhD students.

At first, we presented the landscape of the instance to the contestant.
Then, we introduced them the goal, i.e., covering as much space as possible and the resources to reach that goal, i.e., the four radars and their corresponding parameters.

Participants has 30 minutes to solve the problem, and they could evaluate as many solutions as they wanted within this time frame. For comparison, the average algorithm wallclock time for one run of $2{,}500$ function evaluations is around five minutes. 
For each contestant, we recorded the best value found and the improvements done during the optimization.

\begin{figure}[t]
    \centering
    \includegraphics[width=0.42\textwidth]{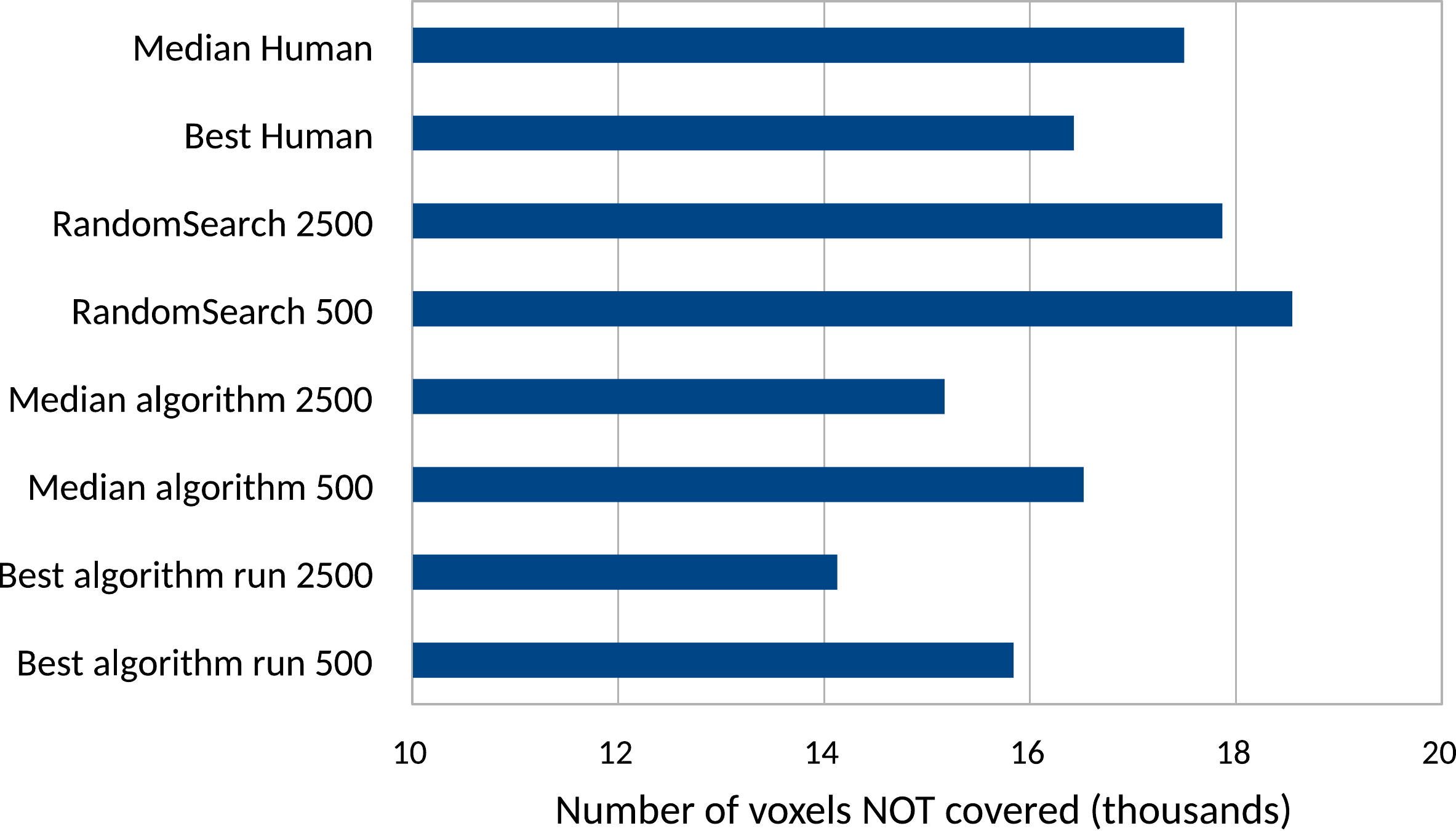}
    \caption{Comparison of performance between solvers and human results.}
    \label{fig:algohuman}
\end{figure}

\textbf{Comparison of Hand-Designed Configurations with Algorithms.} 
The median fitness value obtained by the group is $17{,}495$ voxels not covered. The best result is a configuration that reached fitness value $16{,}425$. 
A few statistics comparing the human-designed results with those of the algorithms are summarized in Fig.~\ref{fig:algohuman}.

In these results, the median performance of the human-designed configuration is around $2\%$ better than random search.
Overall, we see that the optimization algorithms tend to perform much better than the participants of our competition.
The median performance of our participants is $6\%$ worse than the median of best performing algorithm on the budget of $500$ function evaluations and $15\%$ worse on the budget of $2{,}500$ function evaluations.
Moreover, the \emph{canada0} instance seems to be a hard instance for manual optimization, as human performance is close to that of random search.

\section{Automated Algorithm Selection}
\label{sec:auto}

In this section, 
we present how we create our training and testing sets, how we extract landscape features to build our selectors, how we create the mapping between feature data and algorithms performances presented in Sec.~\ref{sec:sbs_unconstrained}, and how we assess the selectors performances.

\subsection{Training and Testing Sets}
\label{sec:lass_train_test}
Training and testing sets are created randomly using all $153$ available problem instances.
The training set is composed of 80\% of the instances, and the remaining 20\% compose the testing set. The instances are selected uniformly at random without replacement. As this random selection can be biased and not representative of the full set of instances, we perform five independent runs to assess the selectors.

\subsection{Feature Computation}
\label{sec:lass_feature_computation}

We describe the two ways that we used to compute problem instance features. Each feature is a numerical value that measures a certain aspect of the problem landscape. The features are computed using \emph{exploratory landscape analysis}~\cite{mersmann_exploratory_2011}. To this end, sample points need to be drawn in the domain and the fitness of the sample points evaluated.
Features are approximated using the pairs of sample points and the associated fitness values.

We experimented with two different ways of extracting features for our problem instances: the traditional one that computes features on the objective function, and a second one that computes the features on the DEM, i.e., without involving the actual radar configuration simulator. 
In both cases, we use the \emph{flacco} package~\cite{flacco2019} to extract the problem instances features. 
We computed all cheap features available in \emph{flacco}, except for the PCA features which do not provide sufficient information when used with non-adaptive sampling (see \cite{renau_expressiveness_2019} for a discussion).
We also excluded NBC features~\cite{kerschke_detecting_2015} as they are not robust to the grid sampling of the DEM.
All experiments are thus based on 33 features in total.

\textbf{Objective Function.} 
\label{sec:obj_func}
For extracting the features on the objective function, we used \sobol low discrepancy sampling~\cite{sobol_distribution_1967}, following a suggestion made in~\cite{RenauDDD20}.  Following common recommendation from the literature~\cite{kerschke_low-budget_2016}, we base the feature computation on $50d=750$ samples.
To compensate for the randomness in the sampling, we perform this feature extraction step $100$ independent times.
The $100$ runs of the features computed on the objective function are aggregated using their \emph{median value}. Hence, for each problem, one vector corresponding to the median values of each feature is used to construct the mapping.

\textbf{Digital Elevation Model (DEM). }
\label{sec:lass_dem}
Building on the idea that the structure of the physical landscape is the factor that has the biggest impact on the objective function, the features can also be computed on the DEM, rather than via the objective function. 

In order to compute features, we need to have an objective function.
In this case, we take the altitude as a function, i.e., we define a function $g(x,y) = z$, where $z$ is the altitude of the point $(x,y)$ in the domain.
Hence, unlike the radar network configuration problem, the dimension of this problem is $d=2$.
On top of that, the domain is discretized, which implies that the number of different altitudes that we can use for the feature computation is at most the number of pixels which is equal to $30\times 30 = 900$.

This approach has two main \emph{benefits}: first, it avoids the rather expensive evaluation of the objective function.
Another advantage, as the domain is discretized, is that we can fully sample the search space.
Doing so, we can remove the randomness of the sampling.

The \emph{drawback} of this approach is the loss of information.
Even though the altitude seems to be the factor that has the biggest impact on the objective function, it is not the only one.
While computing features using the objective function consider all radar parameters (\emph{tilt}, \emph{staring angle}, and internal processing), computing features on DEM has an exclusive focus on the elevation profile of the instance.
Disregarding radar parameters may imply some loss of information on the problem to solve.

In the following, the features computed on the DEM are referred to as DEM features.

\subsection{Building the Mapping between Feature Data and Algorithm Performances}
\label{contrib:lass_mapping}

Using the features extracted by the above-described procedure, we build selectors for the low-budget case and for the large-budget case. To this end, we link the algorithm performance to the feature representation of the instances. We build our model using the median performance of the $30$ independent runs (see Sec.~\ref{sec:sbs_unconstrained}).  

We denote with $S_{\text{radar}}$ the selector built with features computed on the objective function and by $S_{\text{DEM}}$, we denote the selector built with features extracted from the DEM.

As previously done in the literature~\cite{belkhir_per_2017,JankovicD20}, we build the mapping between feature data and algorithm performances using the default scikit-learn Random Forests \emph{regression}~\cite{scikit-learn}.
For each algorithm, we create one Random Forests regression model.

Each regression model learns the performances of the associated algorithm for a given feature vector.
The selector is composed of all the regression models.
When the selector has an unknown feature vector as input, every regression model predicts the performance of its associated algorithm, and the selector then ranks the algorithms by their predicted performance.

\subsection{Definition of the SBS}
\label{sec:def_sbs}
To perform a landscape-aware algorithm selection, we need to split our 153 instances into a training and a testing set (see Sec.~\ref{sec:lass_train_test}).
The splitting can have an impact on the definition of the SBS.
Training splits composed of a majority of mountainous instances may have a different SBS than training splits composed of a majority flat instances.

To visualize the impact of the splits, we create $1{,}000$ independent splits of training instances and record the SBS on both budgets.
On the lower budget of $500$ evaluations, three algorithms were SBS on the splits.
11000000002 was SBS on $503$ splits, DE\_2500\_chile was SBS on $394$ splits, and 01000000000 on $103$ splits.
On the use-cases, we will consider as $\mli{SBS}_{500}$ the 11000000002 CMA-ES variant as this algorithm was more often SBS on the different splits.

Concerning the larger budget, two algorithms were SBS on the splits.
The CMA-ES variant 00100001000 performed best for $986$ splits and DE\_2500\_chile on the remaining $14$ splits.
The $\mli{SBS}_{2{,}500}$ is the same for almost all the different splits.

\subsection{Selector Performances}
\label{sec:select_perf}
We will show in this section the results of our two selectors and how they compare to the virtual best solver (VBS, i.e., the hypothetical optimal selector that always selects the best algorithm for each instance), the SBS, and a common baseline: the vanilla CMA-ES (\emph{00000000000}).

We saw in Sec.~\ref{sec:sbs_unconstrained} that algorithms have similar performances.
This implies that the VBS-SBS gap is small and that the margin to perform landscape-aware algorithm selection is small.

\begin{figure}
    \centering
     \begin{subfigure}[htbp]{0.47\textwidth}
         \caption{$500$ function evaluations}
         \label{fig:selector_reg_one_500}
         \centering
         \includegraphics[trim={0 0 0 1.4cm},clip,width=\textwidth]{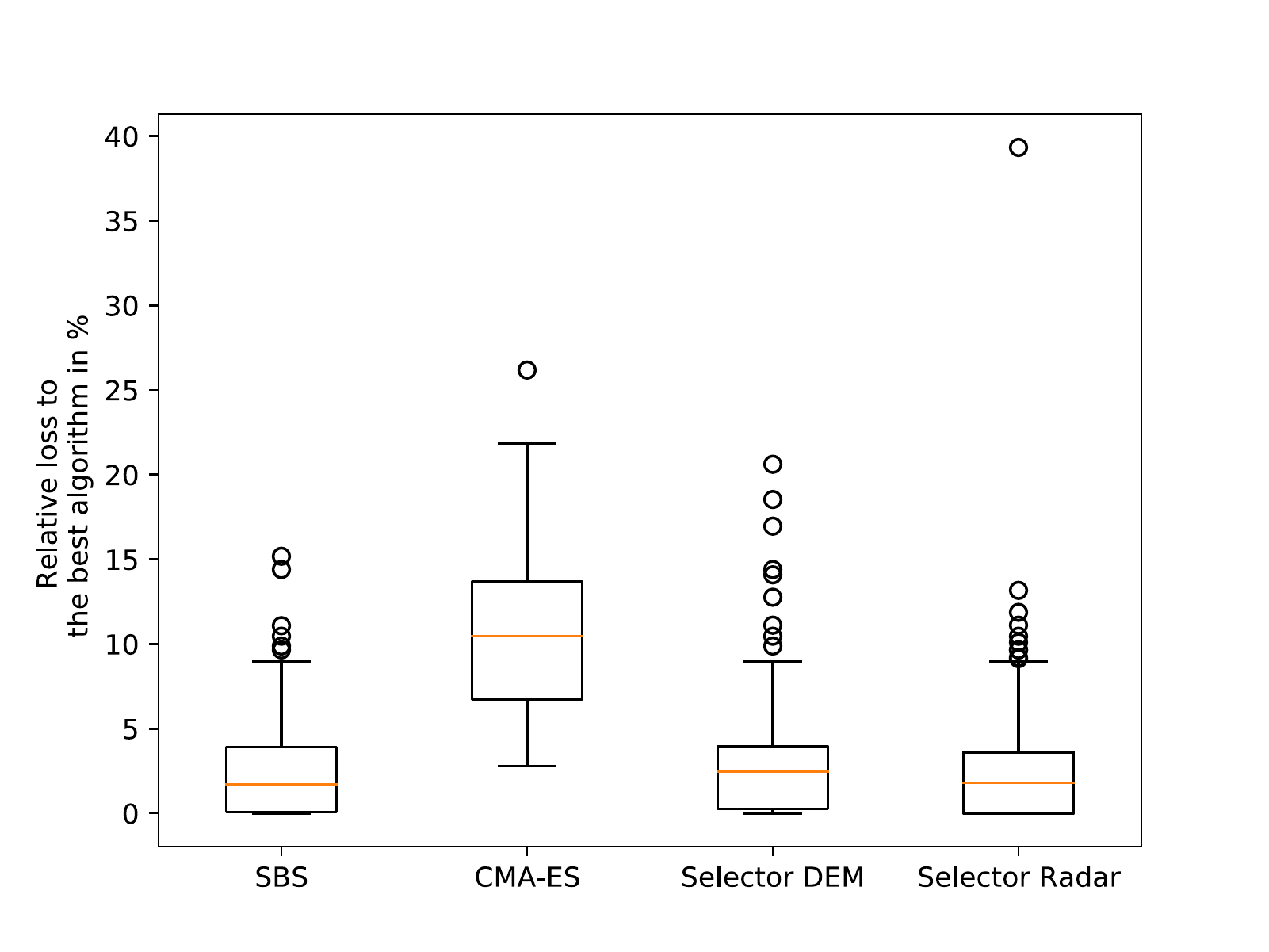}
     \end{subfigure}
     \hfill
     \begin{subfigure}[htbp]{0.47\textwidth}
          \caption{$2{,}500$ function evaluations}
         \label{fig:selector_reg_one_2500}
         \centering
         \includegraphics[trim={0 0 0 1.4cm},clip,width=\textwidth]{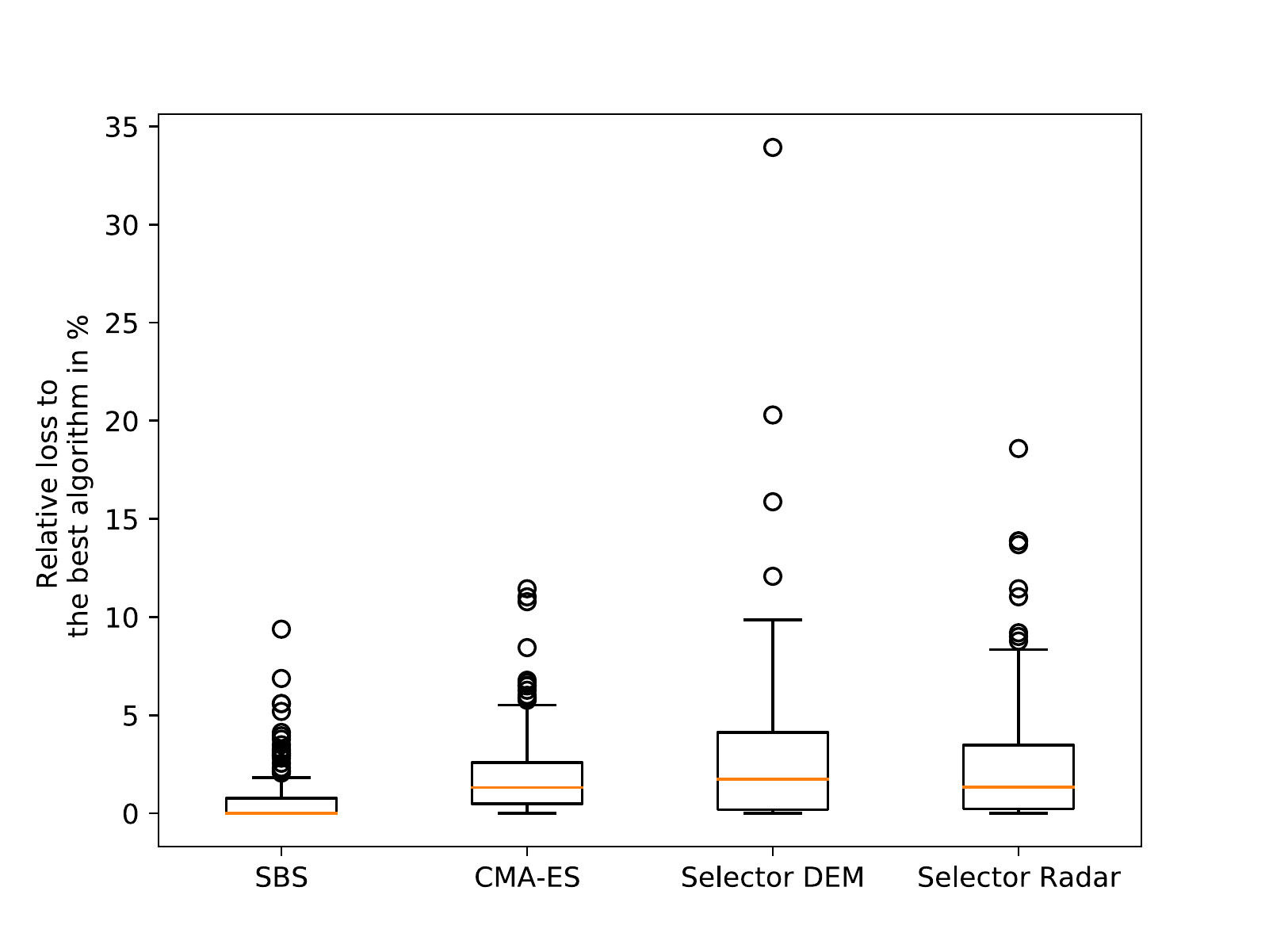}
    \end{subfigure}
     \caption{Relative loss in percentage of the SBS, vanilla CMA-ES, and the selectors with respect to the VBS, assuming that only the best algorithm is proposed.}
     \label{fig:selector_reg_one}
\end{figure}

On the low budget, the selectors outperform the baseline vanilla CMA-ES by around $7\%$ in median performance.
On the larger budget, selectors and vanilla CMA-ES have equivalent performances, i.e., at around $2.5\%$ of the VBS in median performance (see Fig.~\ref{fig:selector_reg_one}).
 
This could be expected from what was mentioned in Sec.~\ref{sec:sbs_unconstrained}. 
Vanilla CMA-ES was not performing well for low budget settings and needed more function evaluations to converge.
On the larger budget, it was one of the best performing algorithm so we expected its performances to be close to those of our selectors.

Overall, the selector perform worse than the SBS for both budgets at $2.6\%$ from the VBS in median
We recall that for this use-case, the median gap between the SBS and the VBS is $0.39\%$ on low budget and $0\%$ for the large budget.

\subsection{Performances of $S_{\text{radar}}$ versus $S_{\text{DEM}}$}

The performances of $S_{\text{DEM}}$ and $S_{\text{radar}}$ are similar.
The median performance of $S_{\text{DEM}}$ is within $0.2\%$ of the median performance of $S_{\text{radar}}$ on both budgets of function evaluations.

Given the small difference in performances, the DEM selector can be preferred to $S_{\text{radar}}$ as its building and using are almost free in computation time.
More precisely, the computation time of the selector $S_{\text{radar}}$ is around $2.5$ minutes when it only takes one or two seconds to use $S_{\text{DEM}}$.
This difference is a consequence of the evaluation of the objective function samples and the feature computation.

\subsection{Discussion}
\label{sec:laas_discussion}

Performances of our selectors are slightly worse than the SBS.
Nevertheless, their performances are quite close to the VBS, i.e., around $2.5\%$ in median for both budgets.

The performance of our selectors may be improved by tuning the Random Forests or by using alternative regression techniques~\cite{JankovicPED21}.
Nevertheless, even if the performances of the selectors increase, the possible gain is relatively low.

Given the performances of the selectors compared to those of the SBS, it would be beneficial to use only the SBS to solve this use-case.
Selectors are not performing badly but the VBS-SBS gap is too small and does not benefit an automated algorithm selection procedure. This may be different for different radar configuration problems or other algorithm portfolios. 
Overall, we consider our results as encouraging for the application of landscape-aware algorithm selection techniques in industrial contexts.  

When there is time to run multiple algorithms, it could be beneficial to run both the SBS and the algorithm recommended by the selector.
For some instances, the algorithm predicted by the selector may have better performances than the SBS.

\section{Conclusion}
\label{sec:conclu}
We have analyzed in this work the performance of 13 metaheuristics on a radar configuration use-case with 4 radars.
For this use-case, we provide binaries~\cite{dataRadar} that allow to reproduce our experiments and that allow to create any benchmark of instances using the model presented.

We have shown that the heuristics outperform human radar configurations, while at the same time requiring much less time than the manual configuration. Our results encourage the use of standard black-box optimization heuristics to aid human experts in the design of radar networks. 

We have also shown that the budget of function evaluations has a major influence on which algorithms to prefer: while elitist versions of the CMA-ES performed well for low budgets, they are outperformed by their elitist counterparts for large budgets. 

We have also shown that a landscape-aware algorithm selection achieves good performance, indicating that the features provided by \emph{flacco} provide useful information to distinguish the problem instances. We have also shown that it suffices to extract features on the DEM. Since this is computationally much cheaper than feature extraction on the objective function, this may pave a way towards real-world deployment of landscape-aware algorithm configuration and selection for radar network configuration.

Extending our pipeline towards more specialized solvers is therefore a straightforward next step of our research. We shall also include other meta-heuristics, e.g., surrogate-based or assisted ones. 

Our experiments are moreover based on a research-style radar configuration model. In industrial contexts, the model complexity often increase with the maturity of the product.
We suspect that the difference between the VBS and the SBS may increase with increasing model complexity, in which case a landscape-aware selection of algorithms may prove more beneficial than in our use-case studied above especially with its ability to better manage a limited evaluations budget. 

\begin{acks}
This work was supported by a public grant as part of the
Investissements d'avenir project, reference ANR-11-LABX-0056-LMH, LabEx LMH, and by the Paris Ile-de-France Region. 
\end{acks}


\newpage
\appendix
\section{Additional Plots}

\begin{figure}[h]
    \centering
     \begin{subfigure}[b]{0.49\textwidth}
                  \caption{\emph{sahara0}}
         \label{fig:sahara_2500}
\centering 
         \includegraphics[trim={0 5cm 0 0},clip,width=\textwidth]{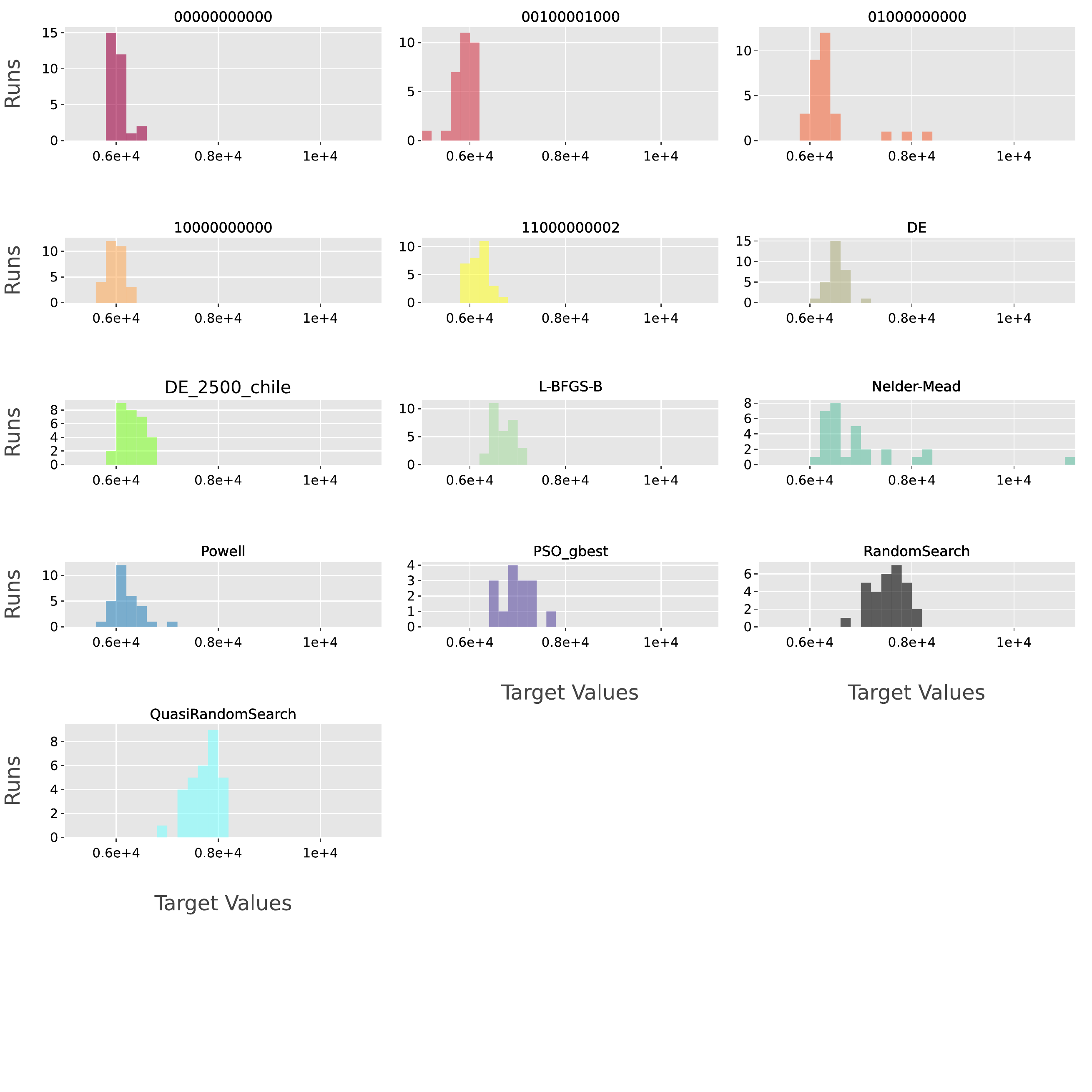}
     \end{subfigure}
     \hfill
     \begin{subfigure}[b]{0.49\textwidth}
         \caption{\emph{nepal3}}
         \label{fig:nepal_2500}
         \centering
         \includegraphics[trim={0 5cm 0 0},clip,width=\textwidth]{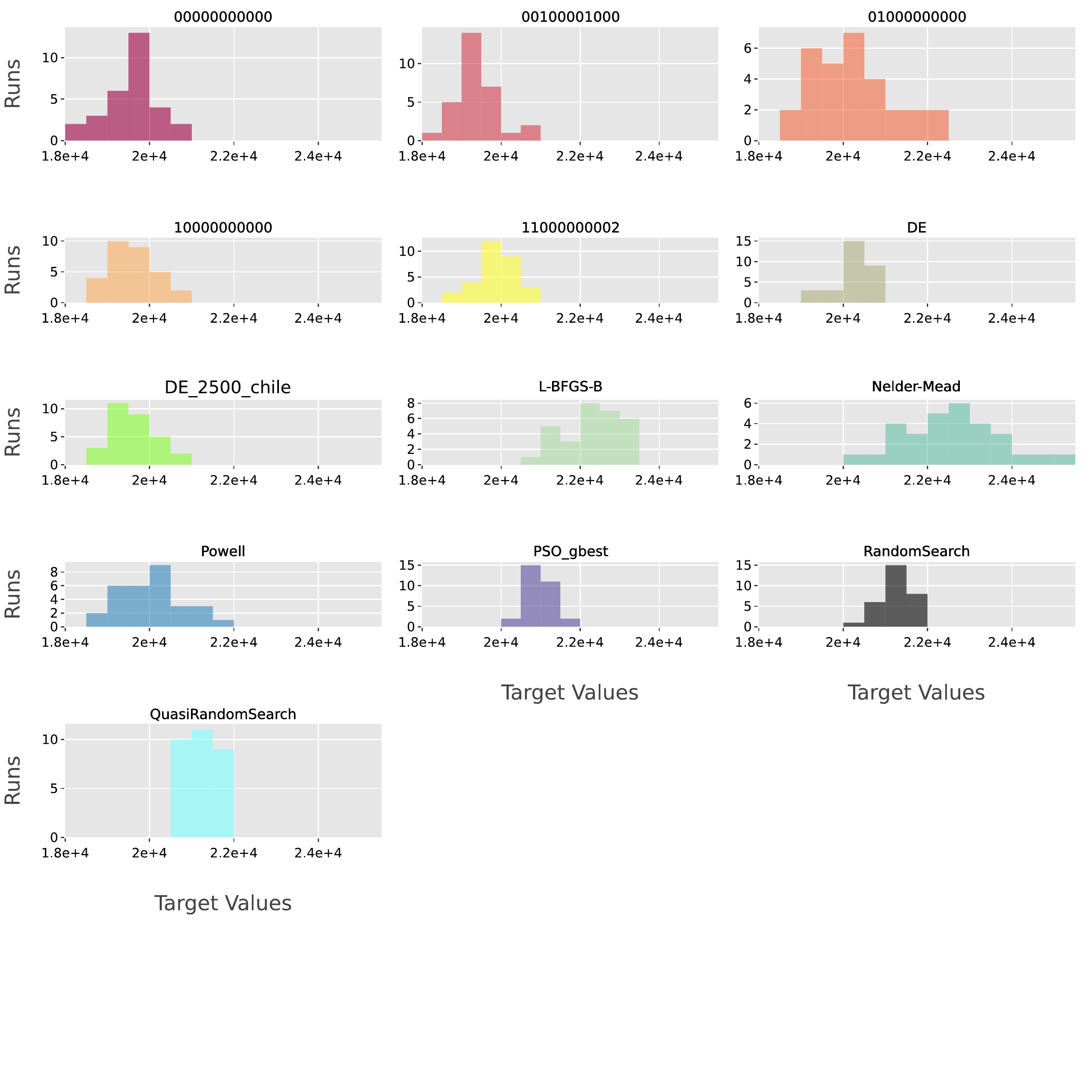}
     \end{subfigure}
     \caption{Histograms of the objective values reached by the 30 runs, for \emph{sahara0} and \emph{nepal3} instances, for the 13 algorithms of our portfolio for $2{,}500$ function evaluations on the unconstrained use-case. }
     \label{fig:histo_2500}
\end{figure}

\begin{figure}[b]
    \centering
     \begin{subfigure}{0.49\textwidth}
         \caption{$500$ function evaluations}
         \label{fig:sbs_500}
         \centering
         \includegraphics[width=\textwidth]{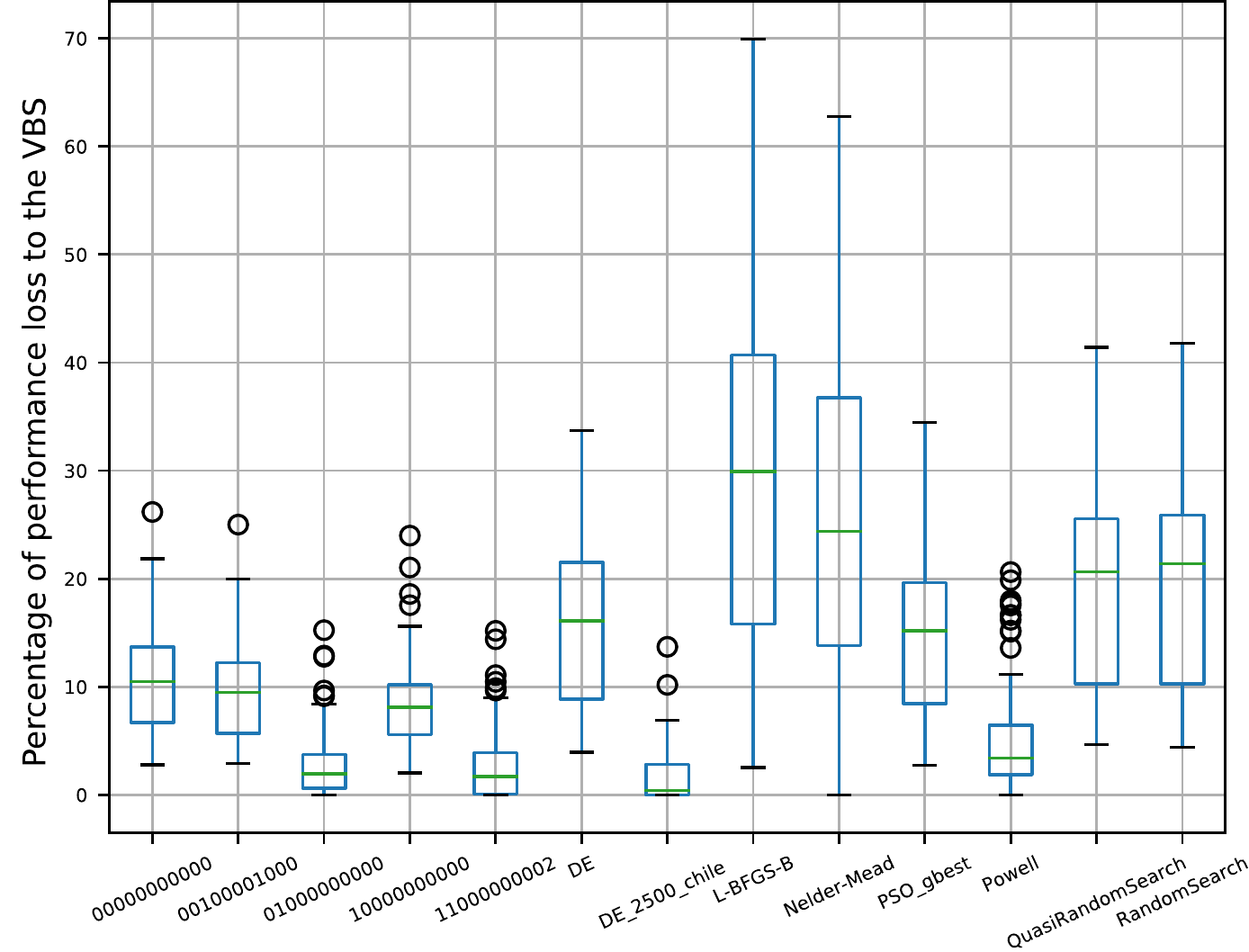}
     \end{subfigure}
     \hfill
     \begin{subfigure}{0.49\textwidth}
         \caption{$2{,}500$ function evaluations}
         \label{fig:sbs_2500}
         \centering
         \includegraphics[width=\textwidth]{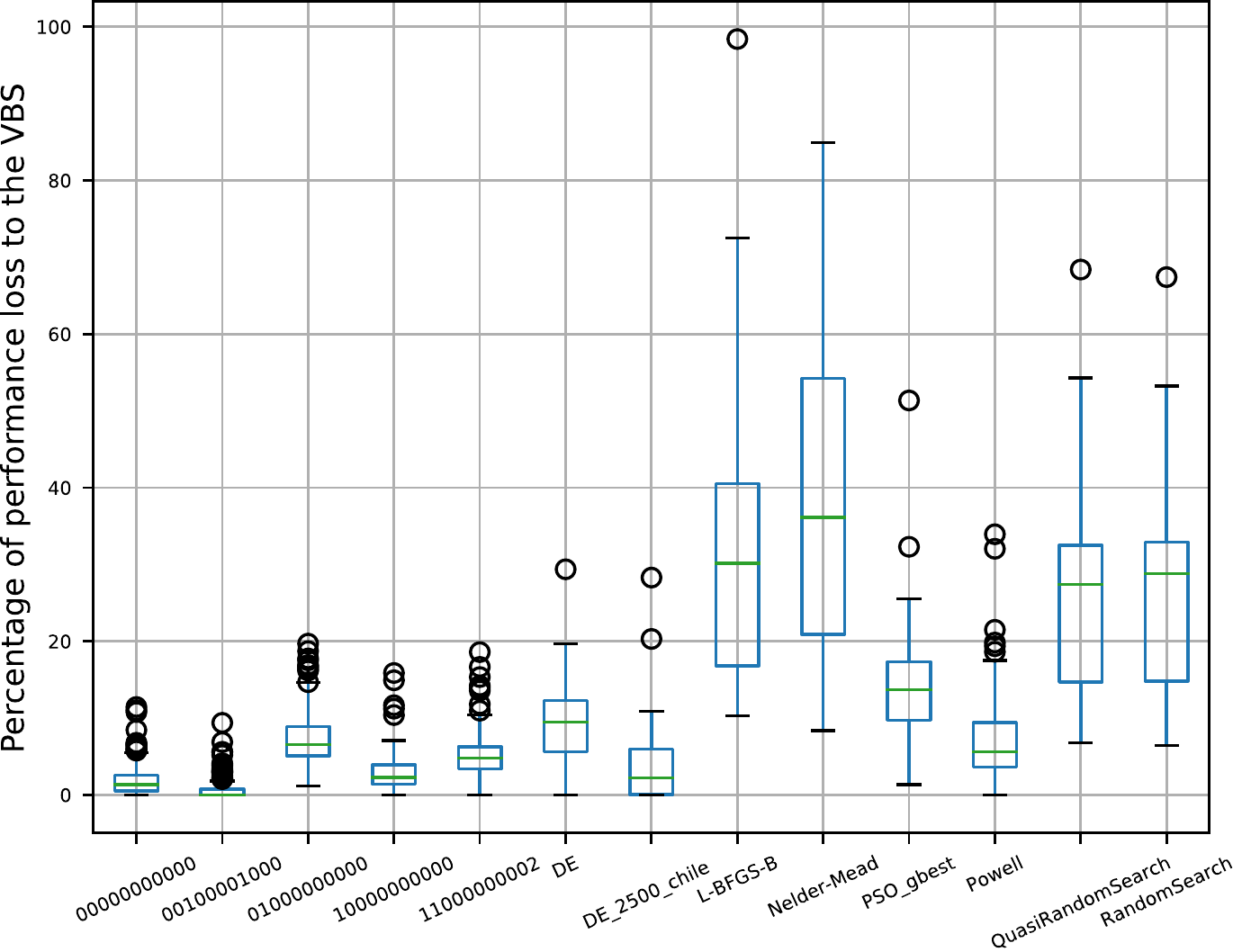}
     \end{subfigure}
     \caption{Boxplots of median performance loss on the best performing solver on all problem instances for low and large budget of function evaluations.}
     \label{fig:sbs}
\end{figure}

\end{document}